# GSVMA: A Genetic-Support Vector Machine-Anova method for CAD diagnosis based on Z-Alizadeh Sani dataset

Javad Hassannataj Joloudari[1,*], Faezeh Azizi[1], Mohammad Ali Nematollahi[2], Roohallah Alizadehsani[3], Edris Hassannataj[4], Amir Mosavi[5,6]

[1]Department of Computer Engineering, Faculty of Engineering, University of Birjand, Birjand, Iran
[2]Department of Computer Sciences, Fasa University, Fasa, Iran
[3]Institute for Intelligent Systems Research and Innovation, Deakin University, Geelong, VIC 3216, Australia
[4]Department of Nursing, School of Nursing and Allied Medical Sciences, Maragheh Faculty of Medical Sciences, Maragheh, Iran
[5]John von Neumann Faculty of Informatics, Obuda University, 1034 Budapest, Hungary
[6]Faculty of Informatics, Technische Universität Dresden, 01069 Dresden, Germany

*Corresponding author: Javad Hassannataj Joloudari (javad.hassannataj@birjand.ac.ir)

**Abstract:** Coronary heart disease (CAD) is one of the crucial reasons for cardiovascular mortality in middle-aged people worldwide. The most typical tool is angiography for diagnosing CAD. The challenges of CAD diagnosis using angiography are costly and have side effects. One of the alternative solutions is the use of machine learning-based patterns for CAD diagnosis. Hence, this paper provides a new hybrid machine learning model called Genetic Support Vector Machine and Analysis of Variance (GSVMA). The ANOVA is known as the kernel function for SVM. The proposed model is performed based on the Z-Alizadeh Sani dataset. A genetic optimization algorithm is used to select crucial features. In addition, SVM with Anova, Linear SVM, and LibSVM with radial basis function methods were applied to classify the dataset. As a result, the GSVMA hybrid method performs better than other methods. This proposed method has the highest accuracy of 89.45% through a 10-fold cross-validation technique with 35 selected features on the Z-Alizadeh Sani dataset. Therefore, the genetic optimization algorithm is very effective for improving accuracy. The computer-aided GSVMA method can be helped clinicians with CAD diagnosis.

**Keyword:**
Coronary heart disease, Genetic Algorithm, Support Vector Machine, Machine Learning.

## 1. Introduction

Cardiovascular disease (CVDs) is one of the most prevalent diseases which cause a lot of deaths in the whole world. As crucial evidence for this fact, one can refer to the cardiovascular diseases fact sheet published by the World Health Organization (WHO), which estimated 17.9 million deaths from CVDs in 2019, representing 32% of all global deaths. Of these deaths, 85% were due to heart attack and stroke [1].

An essential type of CVDs is Coronary Artery Disease (CAD). One of the reasons that made CAD such an essential and stressful disease is the fact that nearly 25% of people who have been diagnosed with CAD died unexpectedly without any prior symptoms [2].

Nowadays, electrocardiogram, cardiac stress test, coronary computed tomographic angiography, and coronary angiogram are some of the prevalent techniques which are used as diagnostic methods for CAD. The downside facts about all these methods are having side effects and imposing high costs on patients and health systems. Hence, today, applying machine learning techniques for diagnosing CAD has become a general tendency. To evaluate the performance of these new techniques, various CAD datasets have been prepared. Among these datasets, the Z-Alizadeh Sani dataset, Cleveland, and Hungarian are public.

In this paper, we applied Support Vector Machine (SVM) with kernel types such as SVM with Analysis of Variance (ANOVA), Linear SVM, and LibSVM with radial basis function to the Z-Alizadeh Sani dataset. This dataset is constructed from 303 patients referred to Shaheed Rajaie Cardiovascular, Medical, and Research Center. For every visitor, 55 features have been considered, all of which are known as signs of CAD according to medical literature [2]. Among the visitors, 216 samples had CAD, and the others were normal. It is worth noting that until now, dozens of studies on the Z-Alizadeh Sani dataset have been published (see, for example, [3-23]). Also, a genetic algorithm as an optimizer is used to select important features in the SVM modeling process. Ultimately, among the proposed methods used in this paper, the genetic optimizer method combined with SVM and ANOVA kernel has the most accuracy of 89.45% on 35 features.

The rest of the paper is organized as follows: Related works are described in Section 2. The proposed methodology is presented in Section 3. Section 4 represents the experimental results and discussion. Section 5 explains the conclusions and future works.

## 2. Related Works

In recent years, studies have been presented using machine learning methods for CAD diagnosis on different datasets. The well-known dataset namely Z-Alizadeh Sani dataset in the field of heart disease is utilized. Hence, we investigated recent works based on the Z-Alizadeh Sani dataset.

In the Qin et al. study [5], several feature selection methods had been implemented on the Z-Alizadeh Sani CHD dataset. The various assessment criteria to evaluate features coupled with a heuristic search strategy and seven classification methods are utilized. They further proposed an ensemble algorithm based on multiple feature selection. According to the authors, the proposed method had better results with a reported accuracy of 93.70% and 95.53% F1-measure.

In Cüvitoğlu and Işik's study [7], an ensemble learner based on the combination of Naïve Bayes, Random Forest, Support Vector Machine, Artificial Neural Networks (ANN), and k-Nearest Neighbor algorithm is developed to diagnose CAD. Also, each of these methods is applied to the dataset separately. The authors performed a t-test for feature selection and reduced the feature space from 54 to 25. Moreover, they implemented PCA to reduce dimensionality further. The best performance between the six methods was reported to be 0.93 AUC which is achieved by the ANN.

Kiliç and Keleş in [9] attempted to select the most convenient features to achieve better performance. They applied the Artificial Bee Colony method on the Z-Alizadeh Sani dataset. The results showed that 16 of 56 features are more meaningful to predict CAD. They reported that a higher accuracy was achieved employing the selected features.

Abdar et al. in [11] proposed a model combining several traditional ML methods using ensemble learning techniques to predict CAD. Also, they employed a feature selection routine based on a genetic algorithm and a filtering method to adjust data. The reported accuracy of this method is 94.66% for Z-Alizadeh Sani and 98.60% for Cleveland CAD datasets.

In [13], Abdar et al. introduced the N2Genetic optimizer, which is a genetic-based algorithm and particle swarm optimization. Using the N2Genetic-nuSVM proposed, they achieved an accuracy of 93.08% and F1-score of 91.51% in Z-Alizadeh Sani dataset for predicting CAD.

In another study, Kolukısa et al. [14] tried two feature selection approaches to extract the most convenient set of features for the Z-Alizadeh Sani Dataset. First, the features were selected based on medical doctor recommendations. Then according to clinically significant findings and Framingham heart study risk factors labeled features. The second method of feature selection was reported to improve the performance of ML algorithms.

A combination of three ensemble learners, random forest, gradient boosting machine, and extreme gradient boosting, form a classifier to predict coronary heart disease (CHD) in the work of Tama et al. [16]. Moreover, a particle swarm optimization-based feature selection model takes the most useful features of data to feed the classifier efficiently. The functionality of the proposed system is verified by having Z-Alizadeh Sani, Statlog, Cleveland, and Hungarian datasets as the input data. The authors claim that the performance of their proposed model outdoes the present methods established on traditional classifier ensembles. They report a 98.13% accuracy, 96.60% F1-score, and 0.98 AUC to classify the Z-Alizadeh Sani dataset.

The effectiveness of Artificial Neural Network and Adaptive Boosting algorithms to predict CAD was tested in the work of Terrada et al. [17]. Data for this study were collected from Z-Alizadeh Sani, Hungarian, UCI repository, and Cleveland datasets, and 17 features were manually selected based on the atherosclerosis risk factors. The results indicated that Artificial Neural Networks show more promising performance over the Adaptive Boosting method.

In [18], Shahid et al. proposed a hybrid Particle swarm optimization-based Extreme learning machine (PSO-ELM) coupled with a Fisher feature selection algorithm to diagnose CAD. Applying the model on the Z-Alizadeh Sani dataset and comparing the results to basic ELM, Shahid declared that their model is a better classifier. The reported accuracy and F1-score for this system are 97.6% and 97.8%, respectively.

In [19], a hybrid algorithm based on Emotional Neural Networks (ENNs) and particle swarm optimization (PSO) is proposed by Shahid and Singh for CAD diagnosis. In addition, they implemented four unique feature selection techniques on the Z-Alizadeh Sani dataset to boost the functionality of the proposed model. According to the authors, their method has a better performance than the PSO-ANFIS model. The F1-score, accuracy, sensitivity, specificity, and precision of the model are reported to be 92.12%, 88.34%, 91.85%, 78.98%, and 92.37%, respectively.

According to Ghiasi et al. [20], only 40 independent parameters of the Z-Alizadeh Sani dataset affect the diagnosis of CAD. The authors apply the classification and regression tree (CART) method for this purpose. They further developed three additional CARD models utilizing 5, 10, and 18 selected features. For the developed model with five features, the reported accuracy is 92.41%. Also, a 77.01% True negative rate and 98.61% True positive value is reported for the model.

Dahal et al. [21] performed Logistic Regression, Random Forest, Support Vector Machine, and K-Nearest Neighbors algorithms for CAD detection on the Z-Alizadeh Sani dataset to determine the most efficient technique. The results indicate that Support Vector Machine has a better performance over other tested methods with 89.47% of accuracy.

Hassannataj et al. [23] have been used the Random trees (RTs) on the 303 samples with 55 features. They have compared the RTs model with SVM, the C5.0 decision tree, and the CHAID decision tree. As a result, using the RTs model, 40 features were ranked with an accuracy of 91.47%, which RTs model has the best performance compared to the other models.

Velusamy et al. [22] developed a heterogeneous ensemble method associated with the output of Support Vector Machine, K-Nearest Neighbors, and Random Forest classifiers based on different ensemble voting approaches. The authors employed average-voting (AVEn), majority-voting (MVEn), and weighted-average voting (WAVEn) to determine which strategy is more successful in predicting CAD. The data for this study were collected from the Z-Alizadeh Sani dataset. The results of this study showed that the WAVEn method is a better voting strategy.

## 3. Proposed Methodology

The proposed methodology has been performed in 3 sub-sections. Section 3.1 describes the Z-Alizadeh Sani dataset. And also, in Section 3.2, data preprocessing will be done. In addition, data classification using SVM with Anova, Linear SVM, and LibSVM with radial basis function and GSVMA methods is described in Section 3.3. The proposed methodology framework is shown in Figure 1.

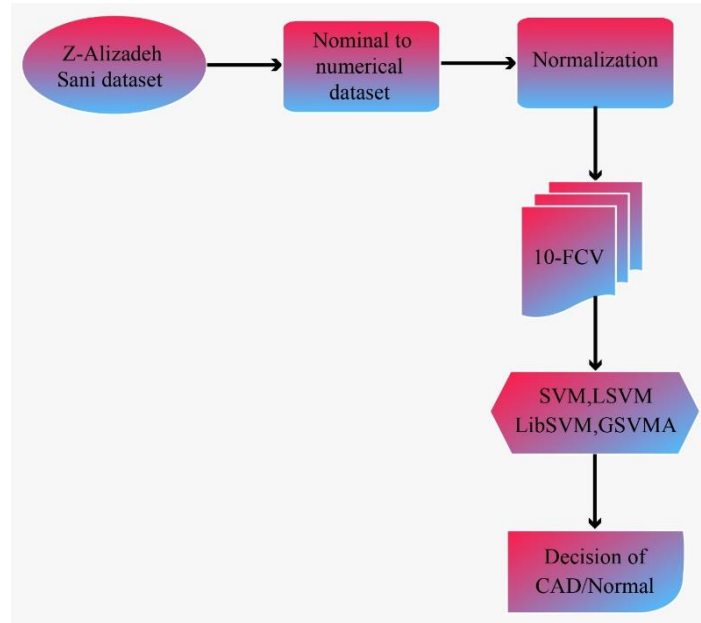

**Figure 1.** The proposed methodology framework.

### 3.1. Z-Alizadeh Sani Dataset

This dataset is one of the most famous datasets used in machine learning for automatic CAD detection. In this dataset, there are 303 records of patients which some of them have CAD. A patient is categorized as a CAD patient if one or more of his/her coronary arteries are stenosis. A coronary artery is categorized as stenosis if its diameter narrowing is greater than or equal to 50% [24]. Accordingly, 216 patients had CAD, and the dataset contains 88 patients with the normal situation on the Z-Alizadeh Sani. Each record in this dataset has 55 features that can be used as indicators of CAD for a patient. These features are grouped into four categories include demographic, symptom and examination, laboratory and echo, and ECG features, which are explained in Table 1.

**Table 1.** Description of the Z-Alizadeh-Sani dataset.

| Feature Type | Feature Name | Measurement | |
|---|---|---|---|
| | | Mean | Std. Deviation |
| Demographic | Age | 58.90 | 10.39 |
| Demographic | Weight | 73.83 | 11.99 |
| Demographic | Length | 164.72 | 9.33 |
| Demographic | Sex | --- | --- |
| Demographic | BMI (body mass index $Kb/m^2$) | 27.25 | 4.1 |
| Demographic | DM (diabetes mellitus) | 0.3 | 0.46 |
| Demographic | HTN (hypertension) | 0.6 | 0.49 |
| Demographic | Current smoker | 0.21 | 0.41 |
| Vemographic | Ex-smoker | 0.03 | 0.18 |
| Demographic | FH (family history) | 0.16 | 0.37 |
| Demographic | Obesity | --- | --- |
| Demographic | CRF (chronic renal failure) | --- | --- |
| Demographic | CVA (cerebrovascular accident) | --- | --- |
| Demographic | Airway disease | --- | --- |
| Demographic | Thyroid disease | --- | --- |
| Demographic | CHF (congestive heart failure) | --- | --- |

| Demographic | DPL (dyslipidemia) | --- | --- |
|---|---|---|---|
| Symptom and examination | BP (blood pressure mm Hg) | 129.55 | 18.94 |
| Symptom and examination | PR (pulse rate ppm) | 75.14 | 8.91 |
| Symptom and examination | Edema | 0.04 | 0.2 |
| Symptom and examination | Weak peripheral pulse | --- | --- |
| Symptom and examination | Lung rates | --- | --- |
| Symptom and examination | Systolic murmur | --- | --- |
| Symptom and examination | Diastolic murmur | --- | --- |
| Symptom and examination | Typical chest pain | 0.54 | 0.5 |
| Symptom and examination | Dyspnea | --- | --- |
| Symptom and examination | Function class | 0.66 | 1.03 |
| Symptom and examination | Atypical | --- | --- |
| Symptom and examination | Nonanginal chest pain | --- | --- |
| Symptom and examination | Exertional chest pain | --- | --- |
| Symptom and examination | Low TH Ang (low-threshold angina) | --- | --- |
| ECG | Rhythm | --- | --- |
| ECG | Q wave | 0.05 | 0.22 |
| ECG | ST elevation | 0.05 | 0.21 |
| ECG | ST depression | 0.23 | 0.42 |
| ECG | T inversion | 0.3 | 0.46 |
| ECG | LVH (left ventricular hypertrophy) | --- | --- |
| ECG | Poor R-wave progression | --- | --- |
| Laboratory and echo | FBS (fasting blood sugar mg/dL) | 119.18 | 52.08 |
| Laboratory and echo | Cr (creatine mg/dL) | 1.06 | 0.26 |
| Laboratory and echo | TG (triglyceride mg/dL) | 150.34 | 97.96 |
| Laboratory and echo | LDL (low-density lipoprotein mg/dL) | 104.64 | 35.4 |
| Laboratory and echo | HDL (high-density lipoprotein mg/dL) | 40.23 | 10.56 |
| Laboratory and echo | BUN (blood urea nitrogen mg/dL) | 17.5 | 6.96 |
| Laboratory and echo | ESR (erythrocyte sedimentation rate mm/h) | 19.46 | 15.94 |
| Laboratory and echo | HB (hemoglobin g/dL) | 13.15 | 1.61 |
| Laboratory and echo | K (potassium mEq/lit) | 4.23 | 0.46 |
| Laboratory and echo | Na (sodium mEq/lit) | 141 | 3.81 |
| Laboratory and echo | WBC (white blood cell cells/mL) | 7562.05 | 2413.74 |
| Laboratory and echo | Lymph (lymphocyte %) | 32.4 | 9.97 |
| Laboratory and echo | Neut (neutrophil %) | 60.15 | 10.18 |
| Laboratory and echo | PLT (platelet 1000/mL) | 221.49 | 60.8 |
| Laboratory and echo | EF (ejection fraction %) | 47.23 | 8.93 |
| Laboratory and echo | Region with RWMA | 0.62 | 1.13 |
| Laboratory and echo | VHD (valvular heart disease) | --- | --- |
| Categorical | Target classes: CAD, Normal | --- | --- |

### 3.2. Data Preprocessing

In the data analysis process, preprocessing is required after data gathering. The Z-Alizade-Sani dataset was numerical and string. First, the number of records is transformed from nominal to numerical values such as gender, chronic, renal, failure, cerebrovascular accident, Airway disease, Thyroid disease, congestive heart failure, dyslipidemia, etc. Then the data is normalized between [0, 1] [23]. Significantly, the feature of the gender relation to woman and man is transformed into zero and one, respectively. In general, normalization leads to an increase in the accuracy of the classification models. Furthermore, a 10-Fold Cross-validation (10-FCV) technique [25, 26] for partitioning the dataset was utilized so that the dataset was divided into training (90%) and test (10%) subsets. The 10-FCV process was run ten times which the results of the methods were obtained by averaging every ten times.

### 3.3. Data Classification Methods
#### 3.3.1. SVM

For the first time, the Support vector machine (SVM) model has been developed for data classification in [27-29], which is a great selection when robust predictive power is required. The SVM is a supervised learning model that transforms data to a high dimensional space, i.e., Hilbert space. Then kernel-based methods due to the visions presented by the generalization theory are exploited and the optimization theory is performed [30]. Indeed, SVM is an area parting model, in which the data allocated into the support vectors are based on machine learning and model constructing [31, 32].

In general, the SVM aims to find the best separator line (optimal hyperplane) between the data of the two classes so that it has the most significant possible distance from all the support vectors of the two classes. These classes are partitioned as linear and non-linear statuses [31, 33]. In these statuses, for the SVM is considered that there is a set of training data $x_1, x_2,...,x_3 \in R^n$ with class $y_1 \in \{1,-1\}$ that are binary $(x_i, y_i)$, $(i=1, 2,...n)$, and $n$ represent the number of training data points.

Since, in the current paper, the linear SVM model has been used as one of the proposed methods for CAD diagnosis, we described this model in the following:

### 3.3.1.1. Linear SVM

The linear kernel is the most common kernel function for Linear SVM (LSVM) [34]. The LSVM model generates an optimized hyperplane that discrete the data points of the two classes.
For an LSVM, a decision function or separator function is defined as follows:

(1) $$f(x_i) = sign(<w, x_i> + b) = \begin{cases} 1 \text{ if } y_i = 1 \\ -1 \text{ if } y_i = -1 \end{cases}$$

In (1), the $w$ parameter represents the weight of inertia, $b$ is the width of the origin point, in which $w \in Z$ and $b \in R$. Based on the LSVM model, an optimized hyperplane is shown in Figure 2 [31].

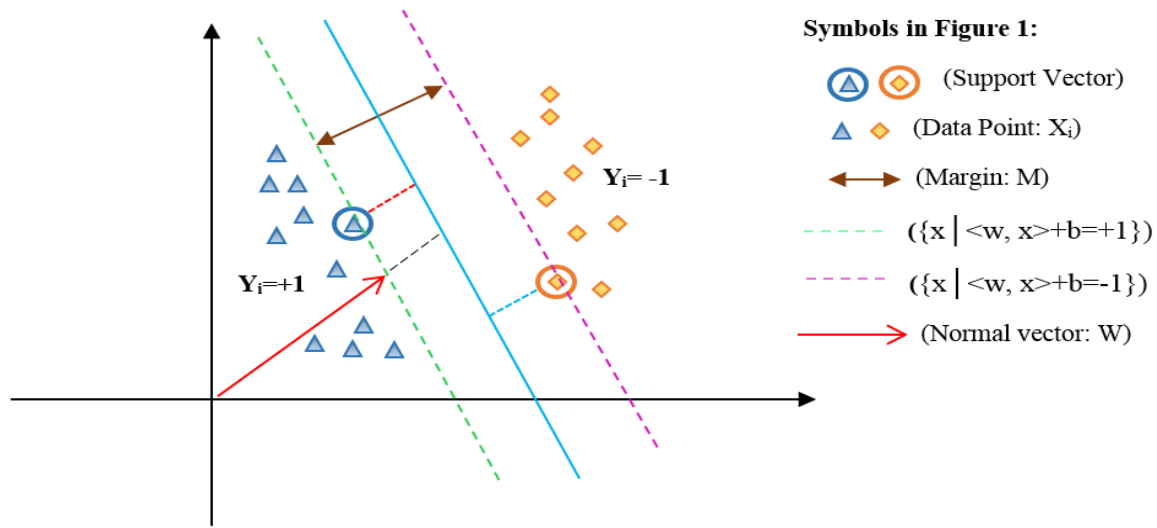

**Figure 2.** Optimized hyperplane for two-dimensional space.

In Figure 1, $x_i$ is the data points of the two classes labeled $y_i = \{1,-1\}$ such that $<w, x> + b = 0$ represents the optimal hyperplane assigned in the average of the other two lines, i.e., $\{x | <w, x> + b = +1\}$ and $\{x | <w, x> + b = -1\}$. Also, $w$ denotes a normal vector for the optimal hyperplane, and $b$ is the offset between the hyperplane and the origin plane. Moreover, the margin $M=w/2$ of the separator is the distance among support vectors. Therefore the maximum margin can be obtained in the form of the following constraint optimization formula (2):

(2) $$\text{Minimize: } \frac{1}{2}|w|^2 \quad \text{subject to: } y_i (w.x+b) -1 \geq 0 \quad \forall i$$

Based on the objective function (2), to solve the optimization problem, the most common approach is to transform it into a dual problem. First, to get the dual form of the problem, the positive Lagrangian coefficients are multiplied by $\alpha_i \geq 0$ and deducted from the objective function, causing in the following equation named a primal problem (Lagrange's initial equation, $L_p$):

(3) $$L_p = \frac{1}{2}|w|^2 - \sum_i \alpha_i (y_i (w.x + b) - 1)$$

To solve formula (3), we are employed the Karush–Kuhn–Tucker (KKT) conditions, which perform an essential role in constraint optimization problems. These conditions state the necessary and adequate conditions for the optimal solution to constraint formulas and must be a derivative of the function regarding the variables equal to zero. By exploiting KKT conditions into $L_p$, has been derived from the $L_p$ relation to $w$ and $b$, and it sets to zero. So, the formulas (4-7) is obtained as follows:

(4) $$\frac{\partial}{\partial w_v} L_p = w_v - \sum_i \alpha_i y_i x_{iv} = 0 \quad v = 1,...,d$$

(5) $$\frac{\partial}{\partial b} L_p = -\sum_i \alpha_i y_i = 0$$

(6) $$\alpha_i(y_i(w.x+b)-1) \geq 0 \quad i=1,...,l \quad \alpha_i \geq 0 \quad \forall_i$$

(7) $$\alpha_i(y_i(w.x+b)-1) \geq 0 \quad \forall_i$$

Consequently, with the assignment of the above formulas into formula (3), formula (8) is gained.

(8) $$Maximize: L_D = \sum_i \alpha_i - \frac{1}{2}\sum_i\sum_j \alpha_i \alpha_j y_i y_j (x_i.x_j)$$

Formula (8) is named the dual problem. Hence, $L_p$ and $L_D$ are both obtained from the same condition. So the optimal problem can be solved by achieving the minimum $L_p$ or the maximum $L_p$, which is the $L_p$ double with the condition $\alpha_i >= 0$.

### 3.3.1.2. Non-linear SVM

If the data are not separable, an LSVM can not be impressive, the non-linear SVM is used so that the data are mapped in a space with higher dimensions. The mapped data from two dimensions space to three dimensions space is illustrated in Figure 3.

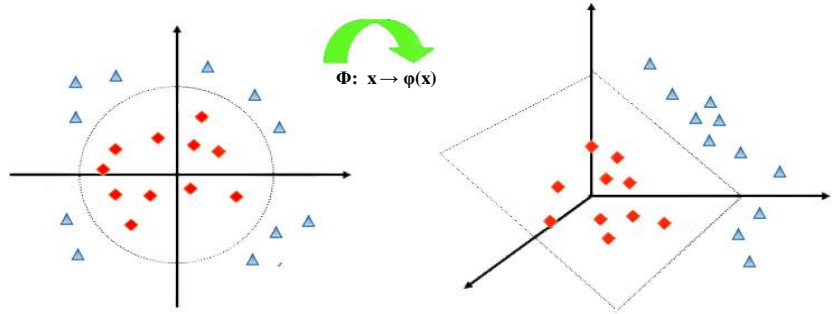

**Figure 3**. The space transmission from two dimensions to three dimensions [31].

Based on Figure 2, first, the data is mapped to a feature space with higher dimensions, i.e., from two dimensions to three dimensions. Then the data can be separated as linear with the indirect mapping instead of the inner product.

So, to solve the feature space problem, a kernel trick is used, which is used a kernel function $k(x_i, x_j)$ in the training stage of the model instead of the inner product $<\phi(x_i), \phi(x_j)>$. If the kernel function has Mercer's condition, the kernel employed is appropriate, and the inseparable data in the mapped space will be distinguishable. Therefore replacing the kernel can build a non-linear model from the linear model. The kernel function $k(x_i,x_j)$ is described in (9):

(9) $$K(x_i, x_j) = (1 + x_i.x_j)^d$$

In (9), *the d* parameter denotes the dimension. Hence, to build a non-linear SVM, a non-linear separator hyperplane is needed, as expressed in (10).

(10) $$Maximize: w(\alpha) = \sum_i^l \alpha_i - \frac{1}{2}\sum_{i=1}^l \sum_{j=1}^l \alpha_i \alpha_j y_i y_j k(x_i, x_j)$$

Subject to $$\sum_{i=1}^l y_i \alpha_i = 0 \quad 0 \leq \alpha_i \leq c \quad i=1,...,l$$

Based on (10), αi and αj are defined as positive lagrangian coefficients, and "c" is a penalty or cost factor for training faults so that the "c" value is randomly chosen by the user. If the value of the "c" increased, more penalties for faults will be made. Then separator function or the decision function is obtained as follows:

(11) $$f(x_i) = sign(<w, x_i> + b) = \sum_{i \in SV} \alpha_i y_i k(x_i, x)$$

In (11), SV is the support-vector. The most common kernel type is the radial basis function (RBF) for SVM.

(12) $$K(x_i, x_j) = \exp(-\frac{\|x_i - x_j\|^2}{2\sigma^2}) = \exp(-\gamma \|x_i - x_j\|^2)$$

According to (12), $\sigma$ is the radial of the kernel function. Also, $\gamma = \frac{1}{26^2}$ represents the kernel parameter. The value of the kernel parameter affects the training rate and the test rate. It should be noted that the efficiency of SVM regarding the accuracy of diagnosis and generalization power is related to the situation of the penalty factor "c" and the kernel parameter "γ" [31].

### 3.3.1.3. Library Support Vector Machine with RBF

The SVM is a binary classifier that it can only classify two classes. A Library for Support Vector Machine (LIBSVM) [35, 36] also supports the multi-class state. The difference between the two-class and multi-class problems regarding training and testing of the data is that larger sets or multi-class conditions may be time-consuming. Indeed, the LIBSVM supports four different kernels by default: linear, polynomial, RBF, and sigmoid kernels, so that the RBF is the most essential tool for SVM and regression classifications developed by C. J. Lin [35]. The RBF kernel is applied in the training phase [34]. The advantage of using the RBF kernel is that it handles the training data to assign specified boundaries. Moreover, the RBF kernel non-linearly maps samples into a higher-dimensional space, and it can handle the training data when the relation between class labels and features is non-linear. The RBF kernel has fewer numerical difficulties than the polynomial kernel, Linear, and Sigmoid. The SVM types are selected through LIBSVM, such as the C-SVC and nu-SVC for classification tasks. Also, the epsilon-SVR and nu-SVR are used for regression tasks, and the one-class SVM is performed for distribution estimation. In this paper, the RBF kernel is selected for SVM as formulated in (12). Moreover, the C-SVC is used to classify data.

### 3.3.1.4. SVM with Analysis of Variance

In general, kernel types are supported by the SVM such as dot, radial, polynomial, neural, Analysis of Variance (ANOVA), epachnenikov, gaussian combination, and multiquadric. The ANOVA kernel is defined by raised to the power "$d$" of summation of $exp(-\gamma (x-y))$ where "$\gamma$" is gamma and "$d$" is degree. The "$\gamma$" and "$d$" are regulated by the kernel gamma and kernel degree parameters, respectively. Indeed, the ANOVA kernel is also a radial basis function kernel. It is said to perform well in multidimensional regression and classification problems [37]. The ANOVA kernel is formulated as follows:

$$(13) \quad k(x, y) = \sum_{k=1}^{n} \exp(-\sigma (x^k - y^k)^2)^d$$

### 3.3.1.5. Genetic Support Vector Machine along with Anova

A genetic algorithm is a method of finding approximate solutions to optimization and search problems. Such an algorithm is a particular type of evolutionary algorithm that uses evolutionary biology techniques such as inheritance and mutation. In a genetic algorithm, to obtain the optimal solution, the appropriate solutions of a generation are combined based on the principle of survival of the most desirable living organisms. In fact, in this algorithm, the solutions to a problem are defined in a chromosome form, which consists of a set of parameters called a Gene. So, Chromosomes are the proposed solutions to the problem of the genetic algorithm.

The most important application of the genetic algorithm is feature selection. Feature selection can be defined as the process of identifying related features and removing unrelated and duplicate features. The feature selection by the genetic algorithm is caused to the better efficiency of the classification methods. Hence, in this paper, we used a genetic algorithm for subset feature selection.

The stages of the genetic algorithm are as follows:

**1) Initial population**

The parameter of the initial population specifies the population size i.e., the number of members per generation. The genetic algorithm starts with a set of chromosomes so that several solutions with different combinations of features are randomly generated. Indeed, the chromosomes of the initial population, which are the initial solutions, are comprised of different combinations of the features. These combinations were randomly comprised for each chromosome and formed the initial solutions to the problem. Hence, in this paper, the purpose of using this algorithm is to find the best subset of the features in the Z-Alizadeh Sani dataset [23]. Using the algorithm, the essential features fed to SVM classification that the best accuracy is obtained for CAD diagnosis. Also, the population size is equal to 50, and the maximum number of generations is set to ten. Therefore the size of each chromosome is related to the number of features including 55 genes for all features. As a result, 35 features are selected using this genetic algorithm conveyed to the SVM classifier.

**2) Determining the fitness function**

The value of each chromosome is determined by the fitness function. This function is used to examine the solutions generated in the initial population. In this paper, maximum fitness has defined infinity.

**3) Selection scheme**

In this stage, based on the fitness criterion, a member of a generation is selected so that members with more compatibility have more probability of making the next generation. The selection schemes such as Uniform, Roulette Wheel (RW), stochastic universal sampling, Boltzmann, rank, tournament, and non-dominated sorting exist in the genetic algorithm [3]. In the current paper, the RW is applied as a select scheme. Based on the scheme, the member with a higher fitness value has more probability of being selected. This scheme is one of the weighted random selection schemes. The probability of choosing each member is obtained according to the formula (14).

$$P_i = \frac{f_i}{\sum_{k=1}^{N} f_k} \quad (14)$$

$P_i$ indicates the probability of choosing the member, "i" $f_i$ indicates the fitness of member, "i" and $N$, the number of members in the initial population. In this paper, the initial $P$ is equal to 0.5. The higher value of $P_i$ represents that the probability of choosing the chromosome is high. In other words, this chromosome has a better chance to produce the next generation.

### 4) The operation of crossover

After the parent chromosomes are selected by the RW, they must be merged to generate two new children for both parents by the crossover operator. In general, there are three crossover types such as one-point, uniform, and shuffle [3, 38]. Using the kind of one-point crossover, a point on two-parent chromosomes is selected and divided into two parts so that one part of the first parent is replaced by one part of the second parent. The other type of crossover is a shuffle that two points on two-parent chromosomes are randomly selected and divided into three parts. Then one part of the first parent is replaced by one part of the second parent, that the children in three parts are a combination of two parents. Finally, using the uniform type, all the chromosomes points are selected as the merge points. First, a random number between zero and one for each part of the chromosome is generated. If this number is less than a constant value, the genes after that point on the chromosome are moved. In this paper, the shuffle has been assigned as the crossover type. Also, crossover "p" is given 0.75.

### 5) Mutation action

After crossover, the mutation action is one of the essential actions to create a new generation. The mutation action is used to modify a member of the current generation to produce a new member. Due to the mutation action by random, the possibility of reaching a better member and escaping the local optimization can be efficient. In this paper, the probability value of 1.0 has been determined.

When the mutation action is performed, the cycle of the genetic algorithm is terminated due to convergence conditions. The convergence condition is determined based on that the number of generations is assigned ten. Again construction action of the new generation should repeat [3, 38].

## 4. Experimental Results and Discussion

In this section, the results of the classification methods such as of the SVM with Anova Linear SVM LibSVM with RBF GSVMA. Based on Table 2, Accuracy (ACC), Positive Predictive Value (PPV), F-measure, Recall, Specificity, and Area Under the Curve (AUC) had been achieved by the confusion matrix.

Table 2. Confusion matrix for diagnosis of CAD.

| The Actual class | The predicted class | |
|---|---|---|
| | CAD | Normal |
| Positive | True Positive | **False Positive** |
| Negative | False Negative | **True Negative** |

Based on Table 2, the elements of the False Positive (FP), False Negative (FN), True Positive (TP), and True Negative (TN) are described as follows:

- FP: The number of samples predicted to be positive is negative.
- FN: The number of samples predicted to be negative is positive.
- TP: The number of samples predicted to be positive is positive.
- TN: The number of samples predicted to be negative is negative.

Also, the evaluation criteria of the methods were obtained to the following formula (15-19) [25].

$$Specificity = TN/TN + FP \quad (15)$$

$$Accuracy = TP + TN/TP + TN + FP + FN \quad (16)$$

$$precision = TP/TP + FP \quad (17)$$

$$recall = TP/TP + FN \quad (18)$$

$$F - measure = 2 * \frac{precision * recall}{precision + recall} \quad (19)$$

Comparing the performance of the methods, the ACC rates of the SVM with Anova, Linear SVM, and LibSVM with RBF were achieved 85.13%, 86.11%, and 84.78%, respectively. Whereas the ACC rate of the GSVMA method is 89.45% based on 10-FCV. According to the other criteria, the GSVMA method had the highest PPV, F-measure, Recall, and Specificity, and is the best predictive method. Moreover, another crucial criterion used to determine the classification methods is the AUC criterion. The AUC indicates the surface area blue the diagram

Receiver Operating Characteristic (ROC). The AUC of all methods is obtained 100%. The comparison results of the evaluation criteria for methods through the 10-FCV are indicated in Table 3.

**Table 3.** The comparison of the methods based on the Z-Alizadeh Sani dataset in this study.

| Methods | ACC(%) | PPV(%) | F-measure (%) | Recall(%) | Specificity(%) | AUC(%) |
|---|---|---|---|---|---|---|
| SVM with Anova | 85.13 | 80.24 | 72.01 | 69.19 | 91.62 | 100 |
| Linear SVM | 86.11 | 77.21 | 75.55 | 74.75 | 90.71 | 100 |
| LibSVM with RBF | 84.78 | 76.24 | 72.38 | 70.03 | 90.74 | 100 |
| **GSVMA** | **89.45** | **100** | **80.49** | **81.22** | **100** | **100** |

The ROC curve was illustrated for all methods in Figure 4.

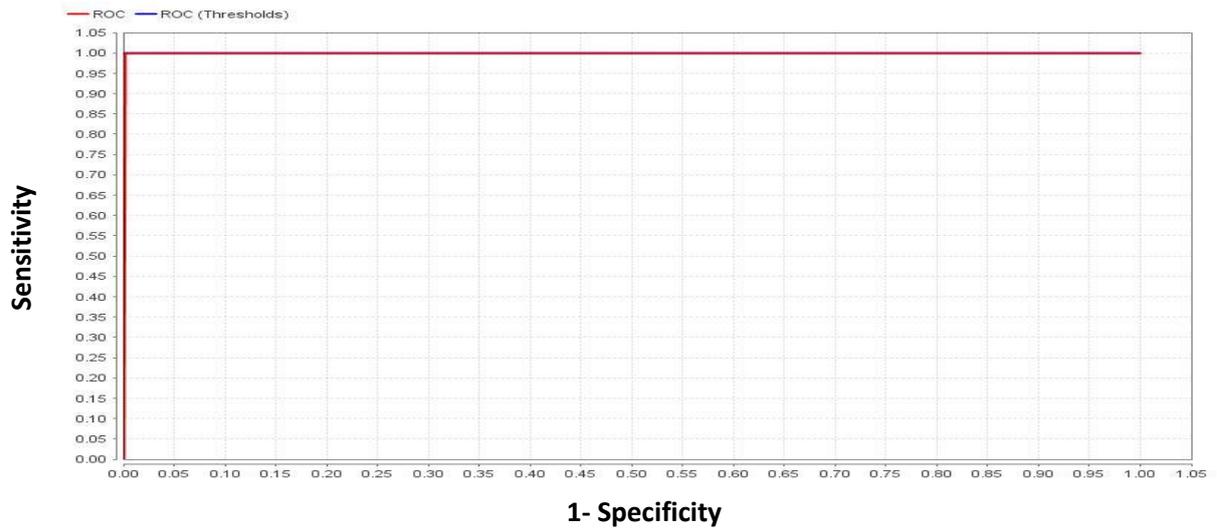

**Figure 4.** ROC diagram for all methods on the Z-Alizadeh Sani dataset.

According to the proposed GSVMA method, the criteria of the ACC, ROC, F-measure, PPV, Recall, and Specificity are illustrated for ten generations in Figures 5-10, respectively.

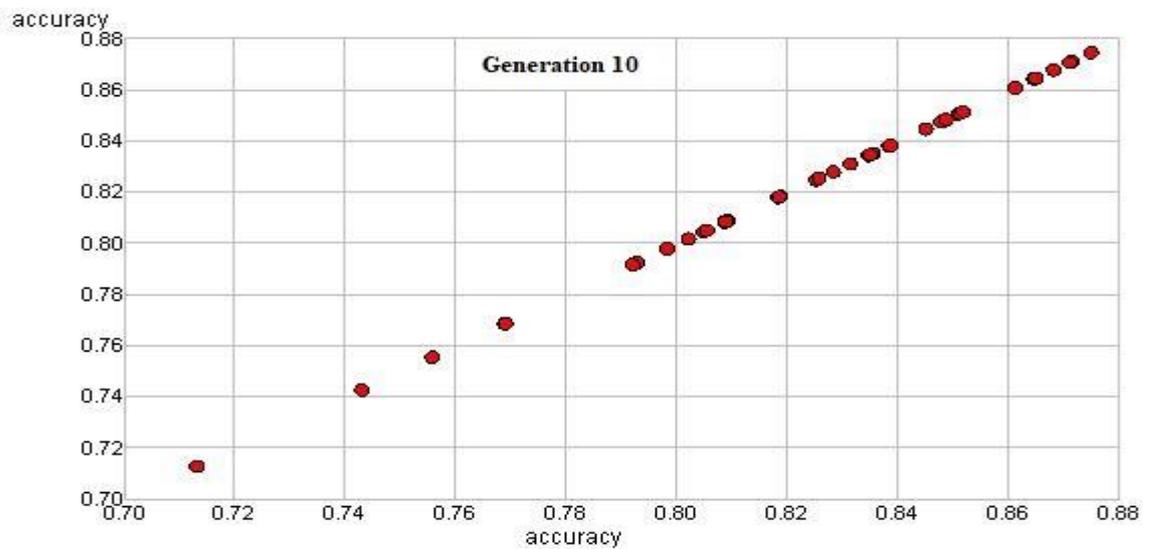

**Figure 5.** ACC diagram for GSVMA method for ten generations.

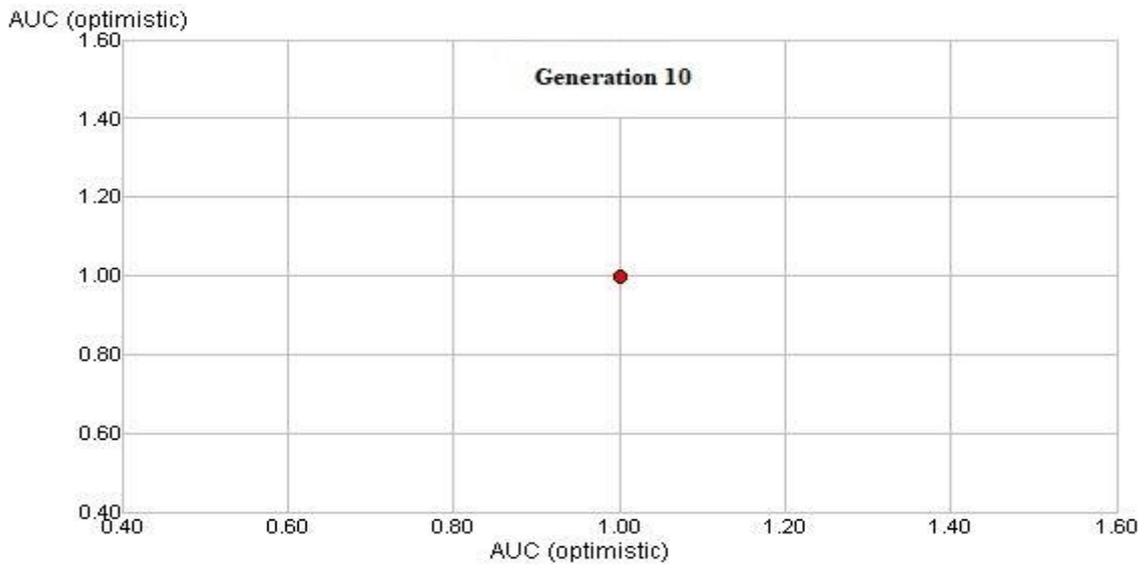

**Figure 6.** ROC diagram for GSVMA method for ten generations.

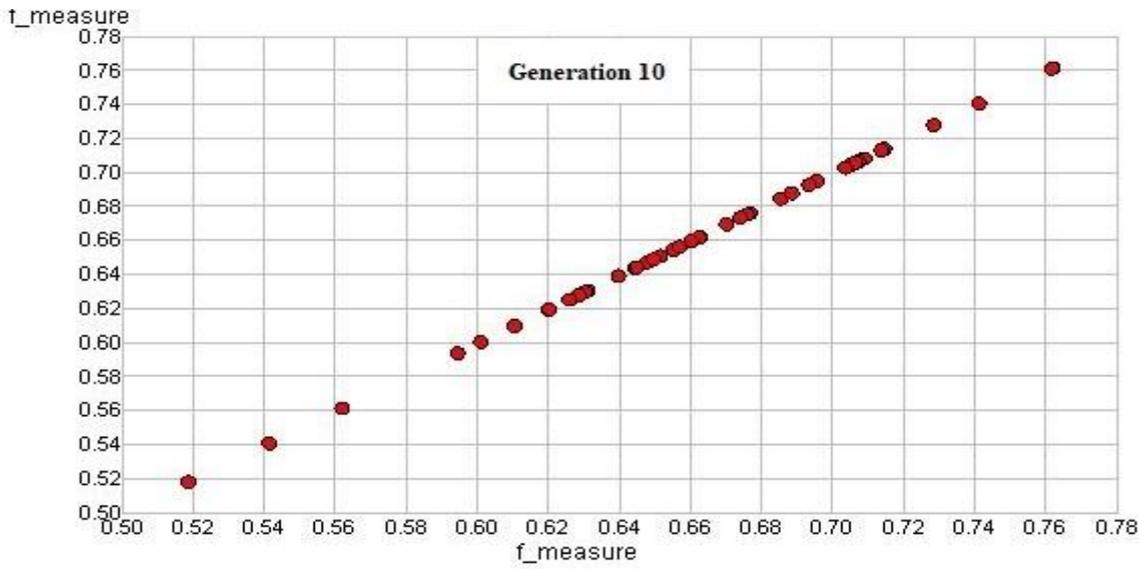

**Figure 7.** F-measure diagram for GSVMA method for ten generations.

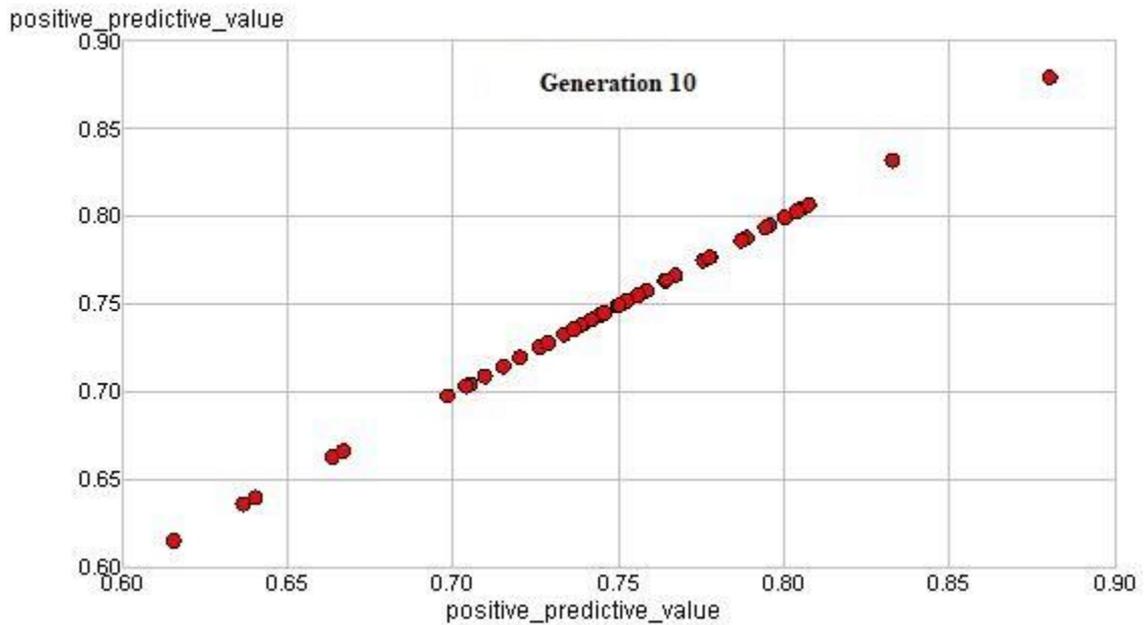

**Figure 8.** PPV diagram for GSVNA method for ten generations.

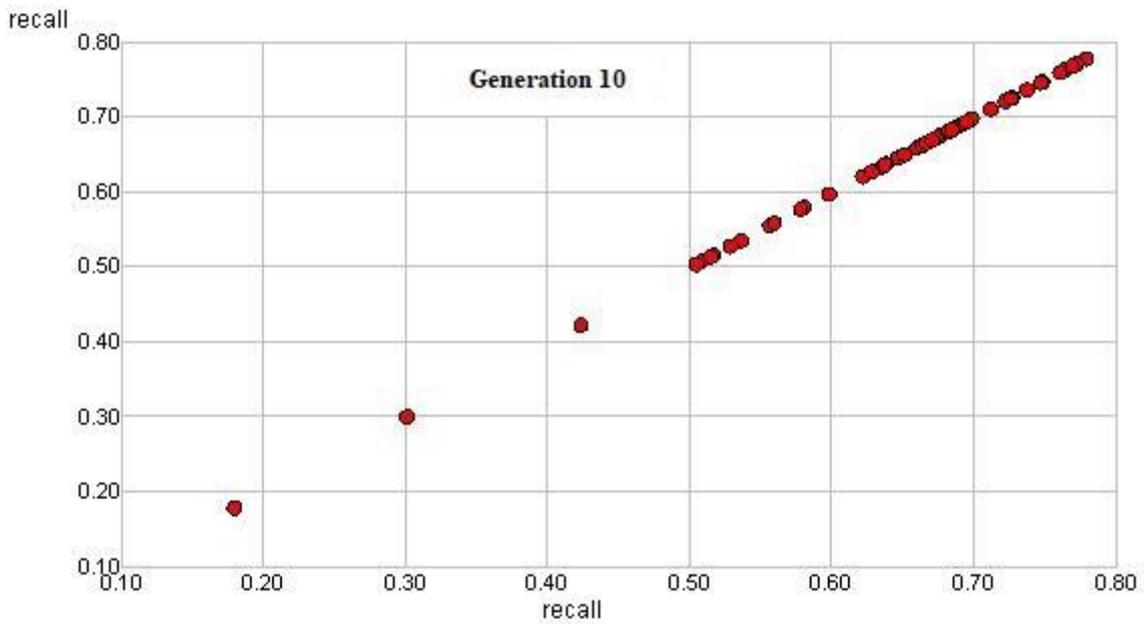

**Figure 9.** Recall diagram for GSVMA method for ten generations.

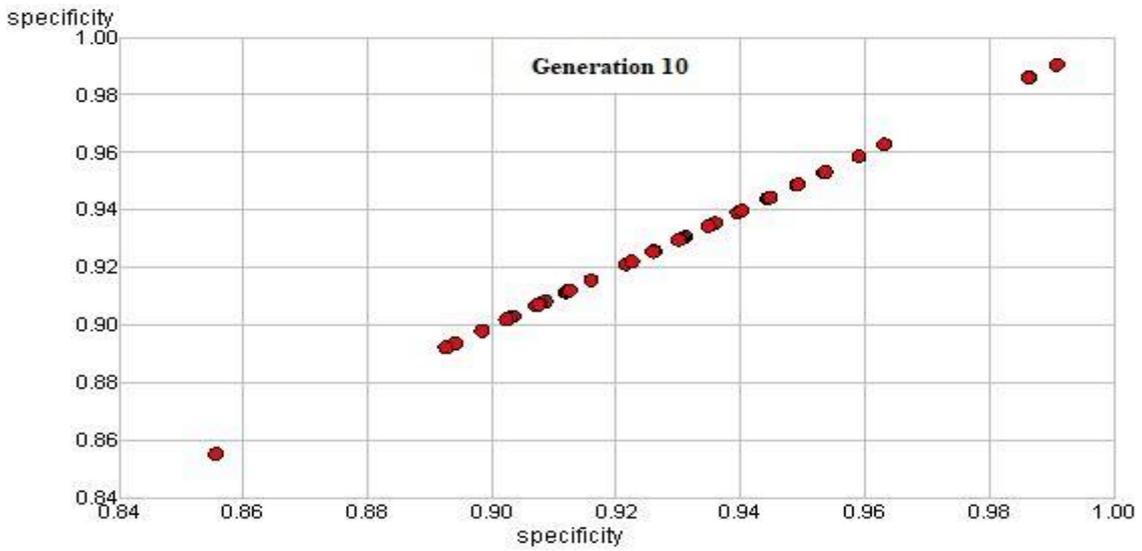

**Figure 10.** Specificity diagram for GSVNA method for ten generations.

Based on Figure 5, the ACC rate is obtained 89.45% through the 10-FCV for ten generations. The AUC of the proposed method is gained 100% for the same generations that the ROC diagram is shown in Figure 6. Also, the F-measure rate has been 80.49% based on Figure 7, and the PPV rate is achieved 100%, as shown in Figure 8. Following, the Recall rate is obtained 81.22%, as illustrated in Figure 9. Finally, based on Figure 10, the Specificity rate has been 100%.

So, the proposed method has the best performance in terms of the mentioned criteria compared to the other methods. Moreover, according to the GSVMA model, the number of features selected through the genetic algorithm is 35 between 55 features, which the crucial features include gender = "male", CRF = "y", CVA = "N", Airway disease = "y", Thyroid disease = "N", CHF = "N", CHF = "Y", Systolic Murmur = "Y", Diastolic Murmur = "N", Diastolic Murmur = "Y", LowTH Ang = "N", LVH = "N", Poor R Progression = "Y", VHD = "mild", VHD = "Severe", VHD = "Moderate", Age, HTN, EX Smoker, FH, PR, Typical Chest Pain, Function Class, Q Wave, St Elevation, T inversion, FBS, TG, LDL, ESR, Lymph, Neut, PLT, EF-TTE, and Region RWMA.

A comparison between the different methods based on the Z-Alizadeh Sani data set is demonstrated in Table 3.

**Table 3.** Comparison between the proposed GSVMA method and the work of other researchers based on the Z-Alizadeh Sani dataset.

| Authors | Techniques | No.Cross Validation | No. Features | ACC (%) |
|---|---|---|---|---|
| Qin et al. [5] | EA-MFS | 10-FCV | 34 | 93.70 |
| Cüvitoğlu et al. [7] | ANN | 10-FCV | 25 | 87.85 |
| Kiliç et al. [9] | ABC | - | 16 | 89.4 |

| Abdar et al. [11] | NE-nu-SVC | 10-FCV | 16 | 94.66 |
| --- | --- | --- | --- | --- |
| Abdar et al. [13] | N2Genetic-nuSVM | 10-FCV | 29 | 93.08 |
| Kolukısa et al. [14] | Ensemble Classifier with FLDA | 10-FCV | 55 | 92.74 |
| Tama et al. [16] | Two-tier ensemble PSO-based feature selection | 10-FCV | 27 | 98.13 |
| Terrada et al. [17] | ANN | - | 17 | 94 |
| Shahid et al. [18] | PSO-ELM | - | 15 | 97.6 |
| Shahid et al. [19] | Hybrid PSO-EmNN | 10-FCV | 22 | 88.34 |
| Ghiasi et al. [20] | CART | 10-FCV | 18 | 100 |
| Dahal et al. [21] | SVM | 10-FCV | 15 | 88.47 |
| Hassannataj et al. [23] | RTs | 10-FCV | 40 | 91.47 |
| Velusamy et al. [22] | WAVEn | 10-FCV | 5 | 98.97 |
| **The Proposed Method** | GSVMA | 10-FCV | 35 | 89.45 |

Due to Table 3, it can be seen that the proposed GSVMA method outperforms other methods regarding accuracy on the 35 features. As a result, using GSVMA is the most informative one about CAD disease.

## 5. Conclusions and Future Works

In this study, the methods such as SVM with Anova, Linear SVM, and LibSVM with radial basis function and GSVMA have been used to diagnose CAD disease. The implementation of the above methods is based on the Z-Alizadeh Sani dataset. In the first step, the Z-Alizadeh Sani dataset has been investigated. This dataset does not contain any missing value and includes 303 samples with 55 features as input and one feature as prediction output. Also, this dataset contains 216 CAD subjects and 87 Non-CAD/Normal Subjects. Next, several approaches were used to preprocess the data. In this stage, the nominal values of features were converted to numeric values. Then the whole samples were normalized, and the 10-FCV method (90% data for training and 10% data for testing) was used for data segmentation. A notable point is that data normalization has led to high diagnostic accuracy. In the third step, data classification has experimented with the above-stated methods. As a result, the proposed GSVMA method with 89.45% accuracy on 35 selected features had the best performance compared to the other methods for CAD diagnosis on the Z-Alizadeh Sani dataset. In future work, different classification methods such as decision tree, random forest, logistic regression, and K-nearest neighbors combined with optimization methods such as genetic algorithm, ant colony algorithm, and gray wolf algorithm can be proposed.